\documentclass[preprint,5p,times,lefttitle,twocolumn]{elsarticle}

\usepackage{amssymb}
\usepackage{amsmath}
\usepackage{subcaption}
\usepackage{multirow}

\journal{Robotics and Autonomous Systems}

\begin{document}

\begin{frontmatter}


\title{Active Informative Planning for UAV-based Weed Mapping using Discrete Gaussian Process Representations}
\author[lincoln]{Jacob Swindell}
\ead{25105508@students.lincoln.ac.uk}
\author[delft]{Marija Popovi\'{c}}
\author[lincoln]{Riccardo Polvara}
\affiliation[lincoln]{organization={Lincoln Centre for Autonomous Systems (L-CAS)},
            addressline={University of Lincoln},
            city={Lincoln},
            postcode={LN6 7TS},
            country={UK}}
\affiliation[delft]{organization={Faculty of Aerospace Engineering, MAVLab},
            addressline={Delft University of Technology (TU Delft)},
            city={Delft},
            postcode={2628 CD},
            country={Netherlands}}




\begin{abstract}
Accurate agricultural weed mapping using unmanned aerial vehicles (UAVs) is crucial for precision farming. While traditional methods rely on rigid, pre-defined flight paths and intensive offline processing, informative path planning (IPP) offers a way to collect data adaptively where it is most needed. Gaussian process (GP) mapping provides a continuous model of weed distribution with built-in uncertainty. However, GPs must be discretised for practical use in autonomous planning. Many discretisation techniques exist, but the impact of discrete representation choice remains poorly understood.
This paper investigates how different discrete GP representations influence both mapping quality and mission-level performance in UAV-based weed mapping. Considering a UAV equipped with a downward-facing camera, we implement a receding-horizon IPP strategy that selects sampling locations based on the map uncertainty, travel cost, and coverage penalties. We investigate multiple discretisation strategies for representing the GP posterior and use their induced map partitions to generate candidate viewpoints for planning.
Experiments on real-world weed distributions show that representation choice significantly affects exploration behaviour and efficiency. Overall, our results demonstrate that discretisation is not only a representational detail but a key design choice that shapes planning dynamics, coverage efficiency, and computational load in online UAV weed mapping.

\end{abstract}



\begin{keyword}
Aerial Systems: Perception and Autonomy \sep Robotics and Automation in Agriculture and Forestry \sep Field Robots \sep Computer Vision for Agriculture


\end{keyword}

\end{frontmatter}



\section{Introduction}
\label{sec:intro}

\begin{figure}[!ht] 
    \centering
    \includegraphics[width=\columnwidth]{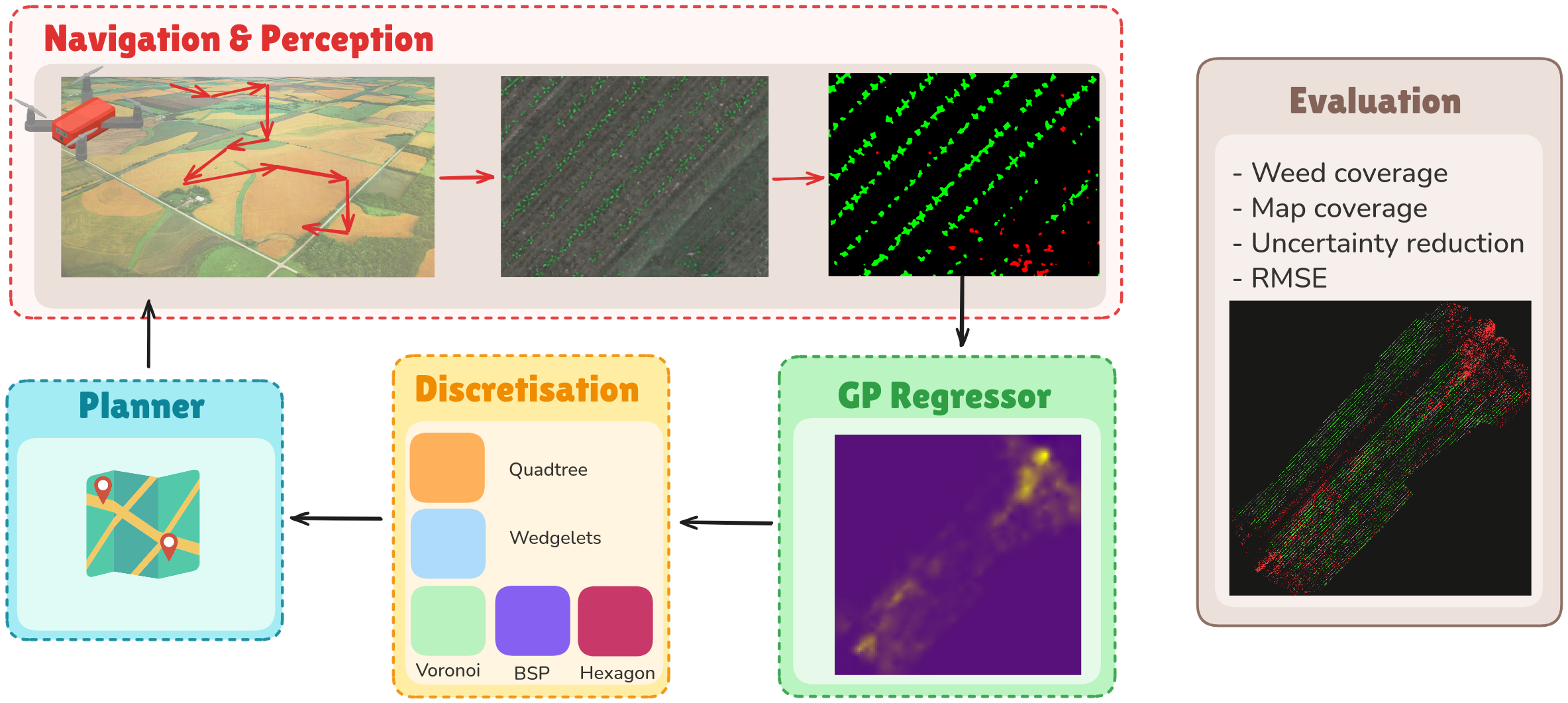} 
    \caption{Methodological overview of the adaptive weed mapping pipeline. A UAV incrementally acquires observations that update a Gaussian process model of weed distribution. The continuous field is discretised using different spatial representations, which generate candidate viewpoints that inform a planning module for autonomous exploration. Evaluation is performed on a fixed grid, allowing representation-independent comparison in terms of coverage, uncertainty reduction, and reconstruction accuracy.}
    \label{fig:pipeline} 
\end{figure}

Accurate weed mapping is a key requirement for precision agriculture, enabling targeted interventions such as site-specific pesticide application. Unmanned aerial vehicles (UAVs) are widely used for this task because of their flexibility, low cost, and high spatial resolution. However, most UAV-based weed mapping pipelines rely on fixed survey patterns, e.g. coverage path planning, and offline orthomosaic reconstruction. While effective for complete field coverage, these approaches typically involve computationally heavy offline processing. Furthermore, they are inefficient when weed distributions are spatially heterogeneous, as sensing effort is distributed uniformly across the field regardless of weed information content.

Gaussian process (GP) mapping~\cite{jin_adaptive-resolution_2022} provides an alternative representation by modelling weed distribution as a continuous spatial field with associated map uncertainty. Unlike orthomosaics, GPs do not require dense, overlapping coverage and can naturally support adaptive data collection. This makes them well-suited for informative path planning (IPP), where a UAV actively selects sensing locations that maximise expected information gain, i.e. weed presence, under resource constraints. As a result, GP-based IPP has the potential to improve mapping efficiency significantly within limited flight time budgets.

A key challenge, however, is that GP models are continuous, while autonomous planners require discrete structures for spatial reasoning and candidate viewpoint generation. Consequently, GP-based weed maps must be discretised into suitable representations. Existing work typically adopts grid maps or quadtrees~\cite{samet_quadtree_1984} by default, without systematically evaluating how alternative discretisation strategies affect preservation of the GP state or downstream planning performance. In particular, the influence of discretised GP representations on online, receding-horizon IPP for UAV-based field mapping remains underexplored.

This paper addresses this gap by studying how different discrete representations of a 2D GP influence both mapping accuracy and mission-level planning performance. We consider a UAV equipped with a downward-facing camera that incrementally observes weed presence while flying over an agricultural field. The collected measurements are used to update a GP model of the weed distribution, which is then discretised to support autonomous planning. We first compare several discretisation strategies - including quadtrees, wedgelets~\cite{gao_application_2008}, binary space partition (BSP) trees~\cite{radha_binary_1991, salembier_binary_2000}, Voronoi partitions~\cite{aurenhammer1991voronoi}, and hexagonal grids~\cite{birch_rectangular_2007} - based on how well they preserve the underlying GP map. We then embed selected representations within an online, receding-horizon IPP framework, where the discretised map is used to propose candidate sensing locations. The planner repeatedly selects short paths that balance information gain, travel effort, and coverage within a fixed flight time budget.

Our results show that discretisation choice has a significant impact on both representation fidelity and planning behaviour. While quadtrees perform robustly across a wide range of weed distributions and enable faster early gains in coverage and uncertainty reduction, alternative representations can be advantageous for specific distribution characteristics. In the online planning setting, different discretisations induce distinct exploration dynamics, affecting coverage efficiency, convergence behaviour, and computational time allocation between planning and traversal.

The main contributions of this work are:
\begin{itemize}
    \item A systematic comparison of discrete representations for 2D GP-based weed maps in terms of map fidelity and computational efficiency on real-world agricultural data.
    \item An online informative path planning framework that integrates discretised GP maps into autonomous UAV exploration.
    \item An empirical study of how different discretised GP representations affect exploration behaviour, weed coverage, and uncertainty reduction under fixed UAV flight-time budgets.
    \item Practical guidelines for selecting GP discretisation strategies based on weed distribution characteristics.
\end{itemize}

Overall, this work demonstrates that discretisation is not merely an implementation detail, but a key design choice that directly shapes the effectiveness of autonomous UAV-based weed mapping.

\section{Related Work}
\label{sec:related}

\textit{Orthomosaics} from UAV imagery are the dominant representation for aerial field mapping, as they provide spatially complete, high-resolution visual reconstructions. These orthomosaics are typically generated through 2D mosaicking, structure-from-motion pipelines, or simultaneous localisation and mapping techniques~\cite{zhang_aerial_2023}, all of which stitch together multiple overlapping images into a single large-scale map. Such reconstruction pipelines are typically executed offline due to the high computational cost of image alignment, feature matching, and global optimisation, which limits their suitability for adaptive or time-constrained UAV missions. The WeedMap~\cite{sa_weedmap_2018} dataset, which is also used in this work, exemplifies an orthomosaic-based pipeline by combining drone localisation with image registration and feature matching between overlapping images. The overlap requirements vary by application, with examples ranging from 80\% front-lap/30\% side-lap~\cite{pena2015quantifying} to 60\% front-lap/30\% side-lap~\cite{lopez-granados_object-based_2016}, while the WeedMap dataset uses 80\% overlap in both directions. These high overlap requirements constrain flight planning flexibility and enforce largely uniform coverage patterns, irrespective of spatial variability in weed distribution.

\textit{Real-time processing} on UAVs has evolved from traditional offline processing~\cite{castaldi_assessing_2017}. In the past, processing power was limited, and small UAVs could not support the necessary hardware. Recently, embedded devices have become more powerful and can incorporate onboard graphics processing units (GPUs) to enable real-time processing of complex algorithms. Modern systems can leverage real-time processing by, for example, using a ground station laptop for path planning while the UAV uses onboard semantic segmentation~\cite{deng_lightweight_2020}. While real-time processing requires smaller, faster models that may compromise segmentation quality, it has been successfully implemented in various applications, including weed identification in winter wheat~\cite{de_camargo_optimized_2021} and for detecting invasive species such as poisonous hogweed~\cite{menshchikov_real-time_2021}. In this paper, we focus on the mapping and planning stages rather than the perception module itself.

\textit{Gaussian Processes (GPs)} are emerging as an alternative to traditional orthomosaic mapping. Their key advantages include the ability to support online updates under sparse sampling, provide principled uncertainty estimates, and interpolate between observations without requiring dense or uniform coverage. For example, GPs have been used for UAV-based magnetic field mapping \cite{kuevor_fast_2023} using custom kernels and probabilistic sensor models to map the influence of magnetic fields. More recent work introduces methods to create adaptive resolution quadtree map representations using incrementally updated GP models via Bayesian filtering~\cite{jin_adaptive-resolution_2022}. Here, an integral kernel is used to calculate correlation over areas of the map rather than single points, which enables adaptive-resolution mapping in regions of interest. This approach aims to improve computational efficiency and map compactness for real-time UAV interpretation, which is beneficial for reactive path planning. While this work explores multiple quadtree construction strategies, it does not consider alternative discretisation structures, nor does it analyse how representation choice influences downstream planning behaviour. Our paper addresses this gap by systematically investigating alternative discrete mapping structures and analysing how their representational properties affect both map fidelity and online path planning performance.

\textit{Image Abstractions} in computer vision vary in their ability to effectively maintain the underlying spatial structure of the data. In an evaluation~\cite{kassim_hierarchical_2009} of quadtrees, binary space partitioning (BSP) trees, wedgelets, and quad-binary (QB) trees, BSP trees showed superior adaptability for representing shapes like diagonal lines compared to more rigid quadtree structures. BSP tree applications include object tracking and occlusion \cite{caccavale_visual_2003}, collision detection \cite{ganter_dynamic_2021}, and image compression \cite{chopra_improved_2011}, with implementations following either top-down division or bottom-up merging approaches. Bottom-up BSP trees have been used for segmentation by~\cite{salembier_binary_2000} through recursive region merging and in UAV hyperspectral imaging \cite{veganzones_hyperspectral_2014} using watershed segmentation. Top-down approaches have utilised methods like the Hough transform for straight-line region detection \cite{radha_binary_1991} and line parameter computation for region division \cite{radha_image_1996}. Wedgelets, as outlined by \cite{gao_application_2008}, represent a modern evolution of wavelets, particularly excelling in preserving anisotropy in image features. Their effectiveness in compression, especially for animation and cartoons, is demonstrated by \cite{lee_2003_image}, while \cite{willett_fast_2004} shows their utility in denoising medical images affected by Poisson noise. Hexagonal maps are well-suited for connectivity analysis, distance representation and visual clarity~\cite{birch_rectangular_2007} and have been applied in diverse fields, from tracking locust movements \cite{klein_application_2023} to creating semantic embeddings for urban mapping~\cite{wozniak_hex2vec_2021}. Voronoi diagrams have been widely applied in robotics, particularly for task allocation in multi-robot systems~\cite{chen2025distributed} and for path planning~\cite{ebrahimi2025bi}. Beyond robotics, they have been employed in image processing as tessellation-based representations~\cite{ahuja1985image} and as a foundation for generating stippled illustrations~\cite{secord2002weighted}. Their ability to vary spatial resolution based on data density makes them particularly attractive for adaptive exploration.

\textit{UAV Exploration} for information gathering is typically divided into informative path planning (IPP), route planning (RP), and autonomous exploration (AE)~\cite{AnicetodosSantos2022}. IPP seeks to maximise information gain within resource limits, RP emphasises coverage and uncertainty reduction, and AE focuses on building environmental representations. These categories often overlap, with many approaches combining aspects of multiple types. Recent work combining IPP and AE applies GPs to model spatial variability for environmental monitoring, e.g. cyanobacteria in lakes~\cite{Hitz2017}. Hitz et al. globally optimise a B-spline path using the Covariance Matrix Adaptation Evolutionary Strategy (CMA-ES) and then replan in a receding-horizon manner to maximise information gain. Similar pipelines have been developed~\cite{jakkala2024multi, mishra2018online, akemoto2022informative} for target search in 3D environments~\cite{meera2019obstacle} and for weed detection~\cite{Popovic2017}, both employing GPs to model observable phenomena and CMA-ES to optimise polynomial trajectories under flight-time and collision constraints. In the latter, a height-dependent camera model is proposed to account for classification accuracy with altitude, a feature particularly relevant for UAV-based weed mapping using top-down imagery. Other IPP-focused methods employ Bayesian Optimisation to produce continuous, information-rich flight paths~\cite{Marchant2014}, while AE-specific approaches adapt domain heuristics such as variable-speed lawnmower patterns for orthomosaic generation~\cite{Krestenitis2024}. Reinforcement learning has also been explored for field weed detection~\cite{vanEssen2025}, training agents to identify weeds from sensor data and generate spatial maps. However, such deep learning approaches remain difficult to interpret.
These methods demonstrate the effectiveness of GP-based IPP for efficient data acquisition, but do not study how the spatial representation of the GP posterior influences candidate viewpoint generation or exploration behaviour

\textit{Our contribution} addresses this gap by systematically evaluating six discrete representations for real-world GP weed maps: quadtrees, wedgelets, BSP trees using least square error, BSP trees using a region-based approach, hexagonal grids and Voronoi diagrams. In contrast to prior work, which treats discretisation as a fixed implementation detail in the mapping method, we explicitly study how representation choice shapes both mapping quality and downstream planning behaviour. 

\section{Methodology}

This section describes the methodology used to evaluate discrete Gaussian Process (GP) representations for UAV-based weed mapping. Our objective is to determine how spatial discretisation affects both map reconstruction and autonomous planning using real-world data. To achieve this, we follow a two-stage process. First, in Section~\ref{subsec:representation} we evaluate the GP representations independently of the planning algorithm to assess how accurately they approximate the orthomosaic data distribution.
Second, in Section~\ref{subsec:exploration} we integrate these representations into an online, receding-horizon UAV planner. This allows us to decouple the mapping method from the planning logic and measure their specific impacts on weed coverage, uncertainty reduction, and trajectory behaviour under realistic flight-time constraints.

\subsection{Representations}\label{subsec:representation}

\paragraph{Dataset / Data Processing}

The WeedMap~\cite{sa_weedmap_2018} dataset provides 5 semantically segmented orthomosaics of different sugar beet fields in Rheinbach, Germany. The semantic images use the labels Weed (Green), Crop (Red) and Background (Black). The orthomosaics provided by this dataset are used to train our GP, based on which we produce the various discretised representations. Additionally, we use a select orthomosaic as the ground-truth for our IPP system to explore. The ground-truth manually segmented images are used for training, and we focus only on the red pixel labels for weeds.

\paragraph{Continuous Weed Distribution}

The GP requires scalar values that are spatially located via $(x,y)$ coordinates for training. To obtain training data for the GP from the orthomosaic map for representation generation, we use random uniform sampling with an average pooling approach. We scatter random points across the orthomosaic and sample images of $150px\times150px$ centred around those points. From these small cropped regions, the average weed value is calculated. 

Training the GP requires selecting a variogram to model the similarity between data points. Usually, a variogram is trained with data-driven methods to match the underlying distribution as closely as possible~\cite{viana_disentangling_2013}. We consider this to be out of the scope of our paper, instead opting to choose the best-performing model from a set of standard variograms. Table \ref{tbl:variogram_metrics} compares the $Q1$, $Q2$, and $cR$ metrics of how different standard variogram models fit the sampled data.

\begin{table}[htbp]
    \centering
    \caption{Variogram statistics. $Q1$ measures the average error of the model predictions compared to the real data; a value close to 0 indicates very accurate predictions. $Q2$ measures if the model errors are consistent with the errors expected; a value close to 1 indicates a reliably accurate model. $cR$ measures the average magnitude of the errors; a low $cR$ shows the errors are small relative to the data scale.}
    \label{tbl:variogram_metrics}
    \setlength{\tabcolsep}{3pt}
    \small
    \resizebox{\columnwidth}{!}{%
    \begin{tabular}{l|c|c|c|c|c|c}
         & Hole-Effect & Exponential & Spherical & Linear & Power & Gaussian \\
        \hline\hline
        Q1 ($\approx$ 0) & \textbf{0.0102} & 0.0124 & 0.0108         & 0.0154 & 0.0216 & 0.0266 \\ 
        Q2 ($\approx$ 1) & 1.206           & 1.776  & \textbf{1.008} & 0.872  & 0.915  & 1.364 \\ 
        cR ($\approx$ 0) & \textbf{0.0005} & 0.0005 & 0.0011         & 0.0014 & 0.0018 & 0.0024 \\ 
    \end{tabular}%
    }
\end{table}

Although Table \ref{tbl:variogram_metrics} shows that the hole-effect variogram produces the best results, due to the non-standard properties of this variogram, we chose to use the exponential variogram instead. Most variogram models' covariance decreases with greater \emph{lag} distances; however, the hole-effect variogram is unique in that it exhibits an oscillatory behaviour where covariance periodically decreases, then increases with greater lag distances~\cite{pyrcz2003whole}. These oscillations do not map well to naturally occurring distributions such as weeds and can lead to unstable or ill-defined matrices, increasing the risk of numerical errors during training. The exponential model provides a more stable and physically plausible prior for spatial weed patterns.

\paragraph{Discrete Representations}

Next, we briefly describe the steps required to construct each representation. We chose all methods based on their common use in the literature for computer vision~\cite{kassim_hierarchical_2009}. Our contribution is to evaluate these representations on 2D GP weed maps to discover the benefits they provide over commonly-used gridmaps and quadtrees.

\textit{BSP Least Squared Error (LSE) Trees}~\cite{radha_binary_1991} work by recursively dividing an image into two homogeneous regions. We first define a region using the corner points of an $n$-gon and then establish parameter domains based online equations. We sample possible dividing lines, pruning them using the LSE Partitioning Line (LPL) transform thresholds~\cite{radha_fast_1991}, and select the best line based on mean squared error calculations. This recursive process continues until either the maximum depth (9) or an arbitrarily chosen homogeneity criterion ($2\times10^{-4}$) is reached.

\textit{Hexagon maps}~\cite{birch_rectangular_2007} produce a variable-resolution representation based on minimising error. Starting with a uniform grid map, we convert coordinate cells to hexagons. We calculate multi-resolution parents by progressively increasing resolution and computing the average ratio of weed to background values and mean squared error for each hexagon. The final representation is created by selecting hexagons based on error thresholds, reflecting the region size and homogeneity.

\textit{BSP Tree Region}~\cite{salembier_binary_2000} approach constructs BSP trees bottom-up, unlike the top-down LSE method. Beginning with a grid map converted to a 4-adjacency graph, the Kruskal algorithm~\cite{najman_playing_2013} constructs an altitude-ordered binary partition tree by merging nodes based on edge weights. The tree is then pruned by merging subtrees with children nodes that occupy $<10$ pixels, setting the minimum region size for the BSP tree. We calculate the average weed values for the filtered regions and display them to produce the final image representation.

\textit{Wedgelets}~\cite{gao_application_2008} combine aspects of quadtree and line-division approaches. It starts with standard quadtree recursion but adds the step of checking for optimal dividing lines within regions. When a region is not homogeneous, we evaluate various line orientations and apply a threshold to determine whether a line can adequately represent the region. Unlike the BSP LSE approach, wedgelets work with square areas and use threshold-based line selection.

\textit{Voronoi diagrams}~\cite{aurenhammer1991voronoi} are used to partition space into regions defined by proximity to the nearest seed point, producing a tessellation composed of Voronoi cells. In this work, Voronoi diagrams are implemented on the GPU through the jump-flooding algorithm~\cite{rong2006jump}, in combination with Linde–Buzo–Gray stippling~\cite{deussen2017weighted}. This method enables the dynamic subdivision and merging of cells according to the underlying GP map, resulting in denser Voronoi cells in regions of high interest and sparser cells in regions of low interest.

\paragraph{Evaluation Metrics}
\label{subsec:eval_metrics_rep}

To evaluate the information loss introduced by discretisation, we compute the following metrics by comparing each representation against a high-resolution grid approximation of the GP posterior.

\begin{enumerate}
\item \textbf{Structural Similarity Index Measure (SSIM)}: Evaluates image similarity by comparing luminance, contrast, and structure, mimicking human visual perception.
\item \textbf{Hamming Distance (HD)}: A perceptual hash captures the visual features and represents them as a string of letters and numbers. Visually similar images have similar hashes, resulting in smaller Hamming distances.
\item \textbf{Mean Squared Error (MSE)}: Measures pixelwise differences between images, focusing on computational accuracy rather than visual similarity.
\end{enumerate}

Together, these metrics capture visual fidelity, structural preservation, and numerical accuracy. We also measure execution time and memory usage for each representation to assess their efficiency and suitability for real-time UAV deployment. 

\subsection{Exploration Pipeline}\label{subsec:exploration}

To provide context for the detailed components described later, we first outline the IPP pipeline. This system is specifically designed to evaluate how different discrete GP representations influence autonomous exploration and weed mapping.

\paragraph{Pipeline Overview}

The process begins by initialising a GP and deploying the UAV from a predetermined starting location. At each step, a discrete representation of the current GP state is generated to serve as the underlying spatial prior for the planner. During the initial sampling phase, where GP outputs can be unstable due to low sample density, a sparse grid approximation is used as a fallback. Once sufficient data is collected, the planner transitions to using the stable discrete representation, extracting the centroids of all leaf cells to serve as candidate view positions.

These centroids form the nodes of a \textit{Delaunay triangulation}~\cite{lee1980two}, providing a structured search space for the planner. Using a receding-horizon approach with a four-step planning budget, the UAV evaluates potential paths based on a utility function that balances:
\begin{itemize}
    \item Information Gain: Prioritising unexplored or high-value weed clusters;
    \item Efficiency: Penalising revisits and excessive path length.
\end{itemize}

After selecting the highest-utility path, the UAV moves to the next location, samples the environment, and updates the GP. To ensure the representation remains accurate as the mission evolves, the GP is fully retrained and hyperparameters re-optimised every ten samples. 

Critically, the entire pipeline---including GP training, representation generation, and path selection---is subject to a realistic 40-minute flight-time constraint. This allows us to observe not just the accuracy of the maps, but the computational trade-offs: how different discretisation methods affect the UAV's ability to cover the field before the battery is exhausted.

\paragraph{Gaussian Process}

A distinct GP configuration is employed within the planner, differing from the CPU-based GP used in the offline representation evaluation. This other CPU-based implementation is characterised by high accuracy but limited computational efficiency. For online path planning, numerical stability and computational speed are required. For this reason, this GP is implemented using \emph{NumPyro} with GPU acceleration, enabling efficient inference and parameter updates.  

We construct the GP as a smooth spatial field model with well-behaved uncertainty. As the initial field is known, we use a zero-mean prior. In addition, we employ a Matérn 3/2 covariance kernel, with log-normal priors placed on the signal variance $\sigma_f^2$, lengthscale $\ell$, and noise variance $\sigma_n^2$. Standard GP regression equations are applied for posterior prediction and uncertainty estimation. The Matérn kernel, described by Eq.~\ref{eq:matern32}, is selected because of its widespread application in environmental and surface mapping tasks within robotics~\cite{zhu2021online,waarum2025mixtures}, where it is used to encode finite smoothness and spatial correlations appropriate for natural phenomena such as weed distributions. The Matérn 3/2 kernel, in particular, provides a suitable compromise between smooth interpolation and responsiveness to local variation, enabling the planner to both identify and exploit regions of interest, i.e. areas of weed density, effectively.  

\begin{align}
  f &\sim \mathcal{GP}\bigl(0,\ k_{\text{M32}}(\cdot,\cdot)\bigr), \label{eq:gp-prior} \\[0.5em]
  k_{\text{M32}}(x,x') 
    &= \sigma_f^2 \left(1 + \frac{\sqrt{3}\,r}{\ell}\right)
       \exp\!\left(-\frac{\sqrt{3}\,r}{\ell}\right), 
       \quad r = \lVert x - x' \rVert_2. \label{eq:matern32}
\end{align}

Non-zero observation noise variance is used to represent sensor uncertainty, and a small jitter term is added to ensure positive definiteness of the covariance matrix. This conditioning facilitates stable log-determinant computations required for mutual information calculations in the planner. 
Eq.~\ref{eq:KXX} presents the covariance matrix used for training points:

\begin{equation}
  K_{XX} = K_{\text{M32}}(X, X; \sigma_f^2, \ell)
           + (\sigma_n^2 + \varepsilon) I,
  \label{eq:KXX}
\end{equation}
with \(\sigma_n^2\) from the `noise' parameter and \(\varepsilon \approx 10^{-6}\) as jitter. Log-normal priors on the signal variance~\eqref{eq:prior-sigma-f}, lengthscale~\eqref{eq:prior-ell}, and noise variance~\eqref{eq:prior-sigma-n} parameters are employed to enforce positive constraints and to incorporate prior knowledge regarding spatial scale. These priors reflect known properties of the normalised coordinate space (0–1); we use an expected inter-weed spacing based on these scales. The following equations show the initial parameter values selected for our GP setup:

\begin{align}
  \sigma_f^2 &\sim \operatorname{LogNormal}\!\left(0,\, 0.5\right) \label{eq:prior-sigma-f} \\
  \ell       &\sim \operatorname{LogNormal}\!\left(\log 0.05,\, 0.2\right) \label{eq:prior-ell} \\
  \sigma_n^2 &\sim \operatorname{LogNormal}\!\left(-3,\, 0.3\right) \label{eq:prior-sigma-n}
\end{align}

These GP parameters are re-estimated online using the No-U-Turn Sampler (NUTS)~\cite{hoffman2014no} every ten new samples. The use of NUTS yields stable and robust posterior estimates, allowing the GP to produce reliable predictive distributions even when available training data are limited.

\paragraph{Planning Algorithm}

After the GP produces the posterior mean and covariance maps, we leverage these to construct a discrete spatial representation. We evaluate two forms of discretisation within the planner, due to their contrasting trade-offs between computational efficiency and geometric flexibility: Voronoi and quadtree representations. The leaf nodes from each representation are used as candidate sampling positions for the planner, and their influence on overall planning efficiency is compared.

A traversable graph is generated from these candidate positions using Delaunay triangulation. The current UAV position is located on this graph using an index map, which identifies the corresponding leaf node containing the UAV. Candidate paths are then generated from the current  UAV position over a fixed planning horizon.

Each candidate path is evaluated according to a composite utility function comprising three components: an information gain term, a path cost term, and a revisit or coverage penalty term, as in the following Eq.~\ref{eq:utility} :

\begin{equation}
  U(\mathcal{P})
  = I\bigl(f; y \mid \tilde{\mathcal{P}}\bigr)
    - \lambda_{\text{cost}}\, C_{\text{path}}(\mathcal{P})
    - \lambda_{\text{visit}}\, \bar{v}(\mathcal{P})
  \label{eq:utility}
\end{equation}

We calculate the information gain term for a candidate path with \(d\) waypoints. After normalisation, the GP is queried at those waypoints to obtain the posterior covariance matrix \(\Sigma_f \in \mathbb{R}^{d \times d}\). This covariance matrix is used with observation noise variance \(\sigma_n^2\) defined from the GP parameter to calculate the mutual information $I(f; y)$ between the latent function values \(f\) and the noisy observations \(y\) along the path:

\begin{equation}
  I(f; y)
  = \frac{1}{2} \log\det\Bigl( I_d + \sigma_n^{-2} \Sigma_f \Bigr),
  \label{eq:mutual-information}
\end{equation}
where \(I_d\) is the \(d \times d\) identity. This is the utility component that rewards paths visiting high-variance, informative locations under the current GP posterior.

The path cost term of a candidate path $C_{\text{path}}$ is calculated from its polyline length in world coordinates. If the path waypoints are \(\{p_0, p_1, \dots, p_{d-1}\}\) with \(p_i \in \mathbb{R}^2\), the cost is

\begin{equation}
  C_{\text{path}}(p_0, \dots, p_{d-1})
  = \sum_{i=0}^{d-2} \bigl\lVert p_{i+1} - p_i \bigr\rVert_2.
  \label{eq:path-cost}
\end{equation}

This is multiplied by a scalar trade-off parameter \(\lambda_{\text{cost}} = 0.15\) to down-weight long, expensive paths.

The revisit penalty term is calculated by generating a binary coverage mask, which records which map cells have already been sensed/captured by the downward-facing camera footprint. For each candidate path, waypoints are projected onto the mask grid. This computes the fraction of samples along the path that fall in already-covered cells.  With \(\lambda_{\text{visit}} = 400\), the revisit penalty term is defined as \(P_{\text{visit}}(\text{path}) = \lambda_{\text{visit}}\,\bar{v}\). where \(\bar{v} \in [0,1]\) is the mean of the coverage-mask values along the path.

After computing the utility of every candidate path, we select the path with maximum utility ${P}^\star$, but execute only its first waypoint $p_1^\star$ in a receding horizon manner:

\begin{equation}
\mathcal{P}^\star = \arg\max_{\mathcal{P}} U(\mathcal{P}), \quad
\text{execute } p_1^\star.
  \label{eq:select-best-path}
\end{equation}

The UAV will then move to and sample from this location. The process repeats in the next step with updated GP data and coverage.

The planner is assigned a fixed time budget. The total duration consumed by GP inference and training, discrete representation generation, planning computation, and UAV motion execution is subtracted from this budget, assuming a constant velocity model. 
The planning process terminates once the allocated time is fully exhausted, ensuring that comparisons between representations reflect realistic trade-offs between computation and physical exploration.

\paragraph{Evaluation Metrics}
\label{subsec:eval_metrics_path}

The planner is evaluated on a single held-out orthomosaic from the WeedMap dataset. The orthomosaic from the dataset was generated with a Ground Sample Distance (GSD) of 1.04cm per pixel. A low, high‑detail flight altitude of 7m is assumed, consistent with similar agricultural mapping tasks~\cite{li2022yield, huang2018uav}. We define an RGB camera with a $33^\circ$ field of view (FOV), which is applied equally in both the horizontal and vertical directions to produce a 1:1 aspect ratio and a square sampling footprint. This configuration simplifies the evaluation procedure and yields a footprint of approximately $400px\times400px$, equal to an area of about $4m\times4m$.

The footprint is calculated using Eq.~\eqref{eq:footprint_pixels} where \(N\) is the number of pixels per side, \(H\) is altitude, FOV is the full field of view (radians or degrees), and GSD is ground sample distance:
\begin{equation}
  N = \frac{2H \tan\left(\tfrac{\text{FOV}}{2}\right)}{\text{GSD}}
  \label{eq:footprint_pixels}
\end{equation}

At each planning iteration, the UAV is directed to a new location selected by the planner, where an image sample of $400px\times400px$ is extracted from the orthomosaic and appended to the GP model. The following performance metrics are recorded at each step:  

\begin{enumerate}
\item \textbf{Weed coverage}: The percentage of total weed pixels in the orthomosaic that have been observed. 
\item \textbf{Map coverage}: The percentage of total map area covered by the UAV.
\item \textbf{Root Mean Squared Error (RMSE)}: calculated using 5000 random samples from the GP surface compared to the corresponding ground-truth values in the orthomosaic, providing an estimate of total map error.
\item \textbf{Map uncertainty}: The mean of the GP posterior covariance map, indicating the average predictive uncertainty and its reduction over time.
\end{enumerate}

We also measure the total distance travelled, along with the computation times for each stage of the processing pipeline, including GP training and regression, discrete representation generation, and planner computation. This allows us to analyse how representation choice affects the allocation of mission resources between sensing, planning, and motion.

\section{Experimental Evaluation}
\label{sec:exp}

Our key aim is to compare the effectiveness of six different discretised GP representations for UAV field mapping. The experiments are designed to answer two complementary questions: (i) how accurately and efficiently different discretisation strategies approximate the underlying GP posterior, and (ii) how these representational differences translate into mission-level behaviour when embedded within an online exploration pipeline. For the following experiments, we used a receding horizon of four, a total time budget of 2,400 seconds (40 minutes), and a constant-velocity model with a speed of 2m/s. All experiments were conducted on an NVIDIA RTX 4070 GPU, an AMD Ryzen 7 7700 CPU (3.8GHz, 8cores), and 32GB of DDR5 RAM.

\subsection{Representations}
This subsection investigates how discrete representations affect the fidelity of GP weed map approximations and their suitability for real-time applications.
\paragraph{Quantitative Results}

Table \ref{tbl:representation_simularity_metrics} shows the mean and standard deviation of each metric defined in Section \ref{subsec:eval_metrics_rep} for each representation across 7 trials. The table shows that the quadtree performs best on average across the 5 orthomosaic maps for SSIM and HD. However, considering the MSE, the BSP LSE approach performs slightly better than the quadtree and wedgelet representations. Since the orthomosaics tested were captured at a roughly $45^\circ$ tilt, it is possible that the BSP LSE can better represent these diagonal regions due to its ability to divide regions at arbitrary angles. Moreover, the BSP LSE approach has the lowest standard deviation, demonstrating a more consistently well-performing representation.

\begin{table}[ht]
    \caption{Similarity metrics results for each and across all the orthomosaics (0-4). Structural similarity is reported as 1-SSIM (e-04). The results are shown in the format "mean(std dev)" across 10 trials. Best results in bold. Lower is better for all metrics. }
    \centering
    {\small
    \setlength{\tabcolsep}{1.7pt}
    \resizebox{\columnwidth}{!}{%
    \begin{tabular}{c|c|c|c|c|c|c|c}
        & \#Map & Quadtree & Wedgelet & BSP LSE & BSP Region & Hexagon & Voronoi \\
        \hline\hline
        \parbox[t]{8mm}{\multirow{5}{*}{\rotatebox[origin=c]{0}{SSIM}}} 
        & 000 & \textbf{0.23(0.13)} & 0.28(0.14) & 4.69(3.19) & 3.91(5.38) & 0.45(0.17) & 0.29(0.12) \\
        & 001 & 0.20(0.11) & \textbf{0.19(0.12)} & 4.05(1.66) & 1.49(0.28) & 0.51(0.10) & 0.20(0.05) \\
        & 002 & 3.20(6.58) & \textbf{3.04(6.04)} & 10.01(15.01) & 3.71(2.06) & 2.38(2.34) & 2.50(5.75) \\
        & 003 & 1.79(1.54) & 1.84(1.55) & 4.20(1.30) & 3.27(1.32) & 12.10(3.65) & \textbf{0.64(0.18)} \\
        & 004 & 12.44(12.76) & 13.19(12.05) & \textbf{7.27(9.07)} & 34.94(33.76) & 15.39(11.46) & 95.56(166.99) \\
        \hline
        \parbox[t]{5mm}{\multirow{5}{*}{\rotatebox[origin=c]{00}{HD}}} 
        & 000 & \textbf{566(130.16)} & 799.43(162.34) & 1762(99.48) & 1323.71(187.03) & 1163.43(200.55) & 1528.86(93.21) \\
        & 001 & \textbf{660.25(134.05)} & 777(157.72) & 1806.25(95.09) & 1390.25(98.04) & 1428(129.20) & 1744(39.16) \\
        & 002 & \textbf{1055.5(101.09)} & 1257.25(96.52) & 1938(64.33) & 1641(152.60) & 1755.75(111.78) & 1709.43(96.39) \\
        & 003 & \textbf{441.5(46.34)} & 590.5(108.63) & 1705.75(60.86) & 1233.75(66.65) & 1883.75(64.99) & 1604.86(109.76) \\
        & 004 & \textbf{560(127.61)} & 676.86(104.82) & 1681.43(37.70) & 1640(235.53) & 2071.43(110.73) & 1970.29(62.89) \\
        \hline 
        \parbox[t]{7mm}{\multirow{5}{*}{\rotatebox[origin=c]{0}{MSE}}} 
        & 000 & \textbf{101.82(46.11)} & 103.28(45.46) & 102.75(27.11) & 129.08(31.04) & 115.89(32.69) & 119.14(26.66) \\
        & 001 & 110.71(33.24) & 110.96(32.52) & 91.39(26.53) & \textbf{89.85(7.46)} & 111.44(19.65) & 101.13(19.00) \\
        & 002 & \textbf{75.57(5.17)} & 78.03(5.04) & 114.88(26.86) & 107.93(29.85) & 108.93(31.65) & 112.85(20.58) \\
        & 003 & 114(27.62) & 114.47(26.63) & 108.29(38.10) & 123.73(30.25) & 97.85(17.47) & \textbf{97.21(35.04)} \\
        & 004 & 146.52(23.97) & 141.38(21.48) & 126.56(19.07) & 127.14(10.62) & 123.64(17.45) & \textbf{122.71(38.59)} \\
        \hline\hline
        SSIM& \parbox[t]{7mm}{\multirow{3}{*}{\rotatebox[origin=c]{0}{AVG}}} &\textbf{3.57(5.11)} & 3.71(5.43) & 6.04(2.57) & 9.47(14.27) & 6.16(7.06) & 19.84(80.02) \\
        HD &  &\textbf{656.65(236.08)} & 820.21(258.18) & 1778.69(101.49) & 1445.74(186.26) & 1660.47(363.63) & 1711.49(171.71) \\
        MSE &  & 109.73(25.51) & 109.62(22.76) & \textbf{108.78(13.16)} & 115.55(16.60) & 111.55(9.48) & 110.61(29.07) \\
    \end{tabular}
    }
    }
    \label{tbl:representation_simularity_metrics}
\end{table}

\paragraph{Qualitative Results}

\begin{figure*}[!ht]
     \centering
     \begin{subfigure}[t]{0.13\textwidth}
         \centering
         \includegraphics[width=\textwidth]{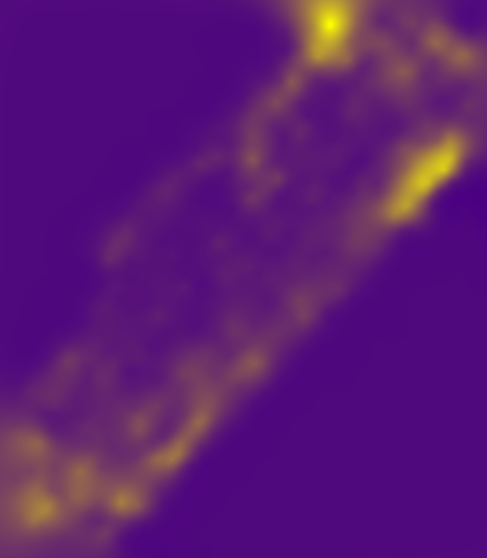}
         \caption{Gridmap} 
         \label{fig:000_gridmap}
     \end{subfigure}
     \begin{subfigure}[t]{0.13\textwidth}
         \centering
         \includegraphics[width=\textwidth]{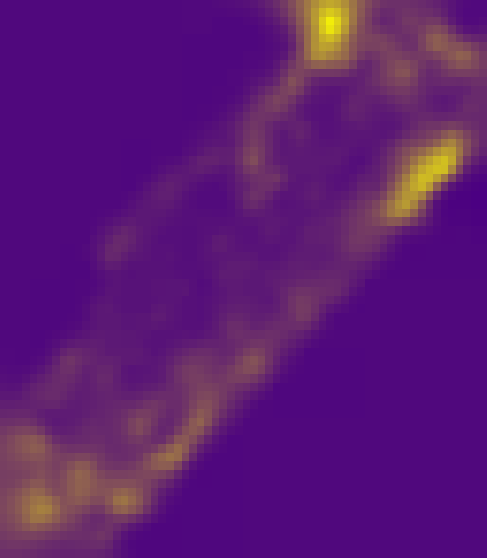}
         \caption{Quadtree}
         \label{fig:000_quadtree}
     \end{subfigure}
     \begin{subfigure}[t]{0.13\textwidth}
         \centering
         \includegraphics[width=\textwidth]{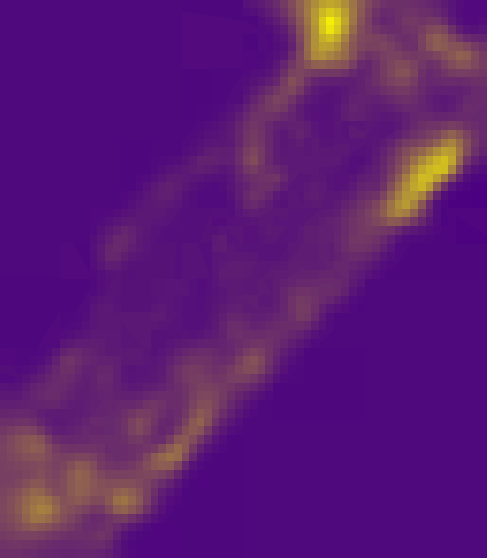}
         \caption{Wedgelet}
         \label{fig:000_wedgelet}
     \end{subfigure}
     \begin{subfigure}[t]{0.13\textwidth}
         \centering
         \includegraphics[width=\textwidth]{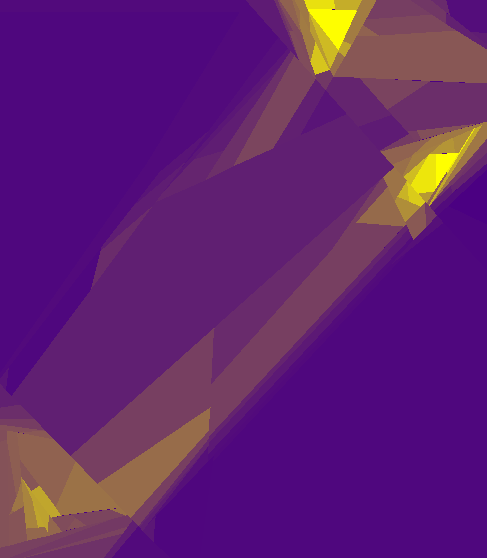}
         \caption{BSP LSE}
         \label{fig:000_bsp}
     \end{subfigure}
     \begin{subfigure}[t]{0.13\textwidth}
         \centering
         \includegraphics[width=\textwidth]{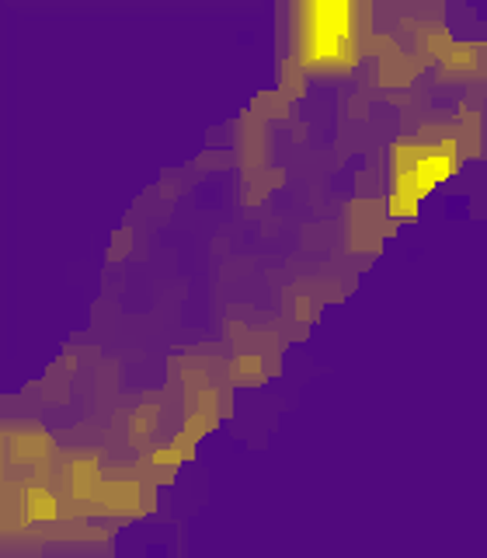}
         \caption{BSP Reg} 
         \label{fig:000_bsp_r}
     \end{subfigure}
     \begin{subfigure}[t]{0.13\textwidth}
         \centering
         \includegraphics[width=\textwidth]{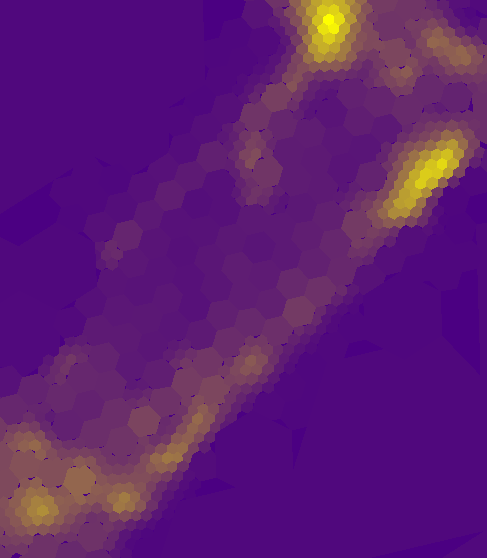}
         \caption{Hexagon}
         \label{fig:000_h3}
     \end{subfigure}
      \begin{subfigure}[t]{0.13\textwidth}
         \centering
         \includegraphics[width=\textwidth]{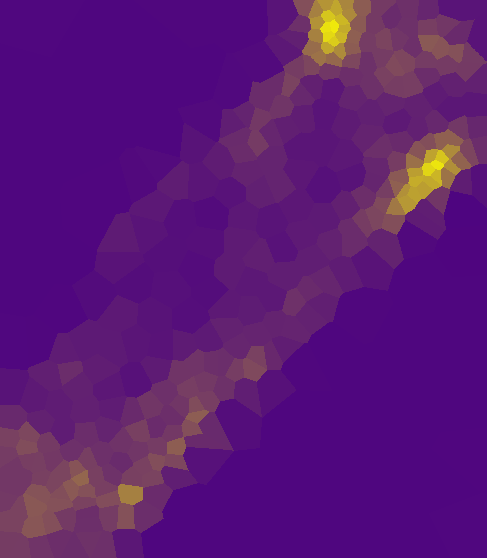}
         \caption{Voronoi}
         \label{fig:000_voro}
     \end{subfigure}
    \caption{Representations for 000\_gt orthomosaic in the WeedMap dataset. Bright spots show a high weed presence, while dark spots show low weed presence. }
    \label{fig:all_orthomosaics}
\end{figure*}

We select the first orthomosaic from the WeedMap dataset (000\_gt.png) to train the GP and create a high-resolution gridmap, from which we compute the different representations. Figure \ref{fig:all_orthomosaics} shows the degree to which they visually capture the high-resolution grid map shown in Figure \ref{fig:000_gridmap}. Bright regions indicate high weed density, while darker regions correspond to sparse or weed-free areas. While BSP LSE (Figure \ref{fig:000_bsp}) appears most abstract, it efficiently represents hotspots with concentrated partitions and fewer partitions in sparse areas. Quadtree (Figure \ref{fig:000_quadtree}) and wedgelet (Figure \ref{fig:000_wedgelet}) are visually similar and the closest to the original grid map, followed by the hexagon representation (Figure \ref{fig:000_h3}). Additionally, the BSP region representation (Figure \ref{fig:000_bsp_r}) has sharp, distinct regional boundaries that accurately show the high weed concentration in that portion of the map. Visually, the Voronoi representation (Figure \ref{fig:000_voro}) resembles the hexagon map, but the Voronoi method produces a more even cell size by comparison. 

\paragraph{Computational and Memory Performance}

\begin{table}[!t]
    \caption{Time to produce representation and memory consumption. Lower is better for all metrics.}
    \centering
    {\small
    \setlength{\tabcolsep}{4pt}
    \resizebox{\columnwidth}{!}{%
    \begin{tabular}{l|c|c|c|c|c|c|c}
        & Quadtree & Wedgelet & BSP LSE & BSP Region & Hexagon & Voronoi & Grid map\\
        \hline\hline
        Time (s) & 0.02(0.01) & 4.13(1.29) & 141(68) & \textbf{0.01(0.00)} & 7.32(2.02) & 4.76(1.26) & N/A \\
        Space (Mb) & 0.77(0.20) & 0.72(0.28) & 0.23(0.08) & 1.02(0.27) & \textbf{0.05(0.03)} & 0.07(0.04) & 11.94(3.14) \\
    \end{tabular}
    }
    }
    \label{tbl:representation_time_space_mean_std_of_mean}
\end{table}

Table \ref{tbl:representation_time_space_mean_std_of_mean} shows the mean and standard deviation of the execution time and memory usage. We compute these metrics by averaging over 7 trials for each of the 5 orthomosaics using the hardware specified in Sec.~\ref{sec:exp}. 
This table additionally shows the memory consumption of the grid map used to train the representations. The BSP region approach is the most computationally efficient, while the hexagon approach consumes the least memory. 

\paragraph{Correlation}

The purpose of our final experiment is to investigate how selected field features (e.g. weed coverage ratio, number of weed patches) of orthomosaics relate to the performance of different partitioning methods. Rather than analysing performance metrics individually, we adopt a composite scoring approach that unifies the three evaluation metrics (Similarity Score, Hamming Distance, and Mean Squared Error) into a single performance score per method. First, we normalise each metric type independently to a 0–1 scale using min-max normalisation, inverting the scales where necessary (SSIM, HD, MSE) so that 1 always represents best performance. We then aggregate the three normalised metrics for each method by computing their mean, yielding a single composite score that balances accuracy across all three evaluation criteria. This consolidation reduces the analytical dimensionality from 18 performance columns (6 methods $\times$ 3 metrics) to 6 method scores, simplifying interpretation while providing a more holistic assessment of each method's performance. We quantify the relationships between field features and these aggregated method scores by computing pairwise Spearman's rank correlations, enabling direct assessment of which partitioning methods are best suited to different weed distribution patterns.

\begin{table}[ht]
\centering
\caption{Spearman's correlation values between features and performance metrics. We report the max and min values along with their associated performance metric.}
\resizebox{\columnwidth}{!}{%
\begin{tabular}{p{6cm} l c @{\hspace{1cm}} l c}
\hline
\textbf{Feature} & \multicolumn{2}{c}{\textbf{Positive Correlation}} & \multicolumn{2}{c}{\textbf{Negative Correlation}} \\
\cline{2-3}\cline{4-5}
& \textbf{Method} & \textbf{Value} & \textbf{Method} & \textbf{Value} \\
\hline
weed\_coverage\_ratio (\%)  & BSP\_LSE & 0.9000 & Quadtree, Wedgelet & -0.5000 \\
num\_weed\_patches          & H3 & 0.9000 & Quadtree, Wedgelet & 0.3000 \\
largest\_patch\_fraction (\%)   & Quadtree, Wedgelet & 0.0000 & BSP\_Region, Voronoi & -0.9000 \\
avg\_patch\_size (px)       & BSP\_LSE & 0.7000 & Quadtree, Wedgelet & -0.8000 \\
patch\_size\_std (px)       & BSP\_LSE & 0.1000 & BSP\_Region, Voronoi & -0.7000 \\
dbscan\_num\_clusters       & H3 & 0.9000 & Quadtree, Wedgelet & 0.3000 \\
dbscan\_avg\_cluster\_size  & BSP\_LSE & 0.7000 & Quadtree, Wedgelet & -0.8000 \\
\hline
\end{tabular}%
}
\label{tab:pos_neg_corr}
\end{table}

The correlations in Table~\ref{tab:pos_neg_corr} (lower scores denote better performance) reveal distinct operating regimes. Coverage-driven features (such as weed\_coverage\_ratio) are strongly positively associated with BSP LSE ($+0.9$) but moderately negatively with Quadtree/Wedgelet ($-0.5$), indicating BSP LSE degrades as coverage intensifies while Quadtree/Wedgelet tend to improve. Fragmentation indicators (num\_weed\_patches, dbscan\_num\_clusters) correlate very positively with H3 ($+0.9$), signalling H3 is ill-suited to highly fragmented fields; Quadtree/Wedgelet show only weak associations ($+0.3$). When a single patch dominates (largest\_patch\_fraction), BSP\_Region and Voronoi exhibit strong negative correlations ($-0.9$), performing best under dominance, whereas Quadtree/Wedgelet are largely insensitive ($0$). Larger average patches favour Quadtree/Wedgelet ($-0.8$) and penalise BSP LSE ($+0.7$); greater patch-size variability favours BSP\_Region/Voronoi ($-0.7$) and has a negligible effect on BSP LSE ($+0.1$). Collectively, these trends recommend Quadtree/Wedgelet for high-coverage fields with sizeable patches, BSP\_Region/Voronoi when dominance or strong heterogeneity is expected, caution against H3 in heavily fragmented settings, and caution against BSP LSE as coverage and patch size increase.

\subsection{Informative Path Planning}
This section evaluates the impact of discrete GP representations on mission-level performance within an online exploration framework, where all planning parameters are held constant to isolate the influence of the spatial representation.

\paragraph{Quantitative Results}

\begin{figure}[ht]
     \centering
      \begin{subfigure}[t]{0.48\columnwidth}
         \centering
         \includegraphics[width=\columnwidth]{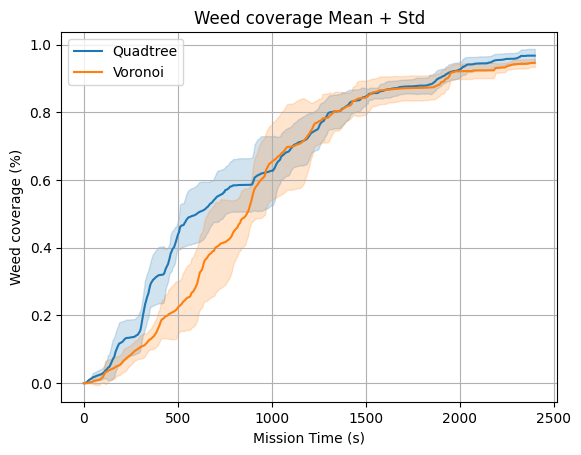}
         \label{fig:planner_weed_coverage}
     \end{subfigure}
     \begin{subfigure}[t]{0.48\columnwidth}
         \centering
         \includegraphics[width=\columnwidth]{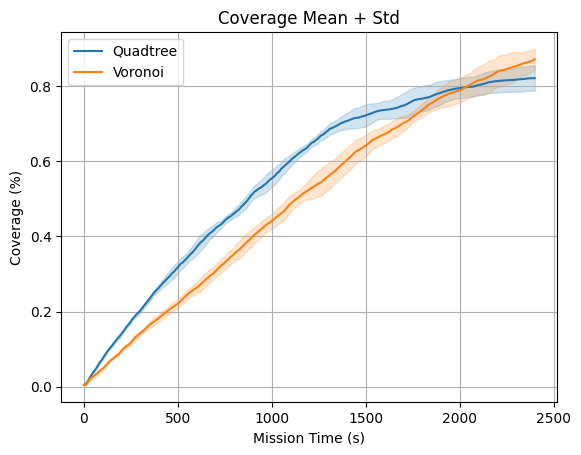}
         \label{fig:planner_coverage}
     \end{subfigure}
     \begin{subfigure}[t]{0.48\columnwidth}
         \centering
         \includegraphics[width=\columnwidth]{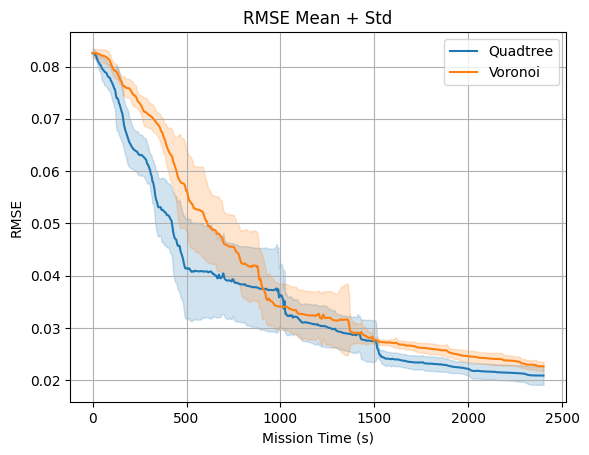}
         \label{fig:planner_rmse}
     \end{subfigure}
     \begin{subfigure}[t]{0.48\columnwidth}
         \centering
         \includegraphics[width=\columnwidth]{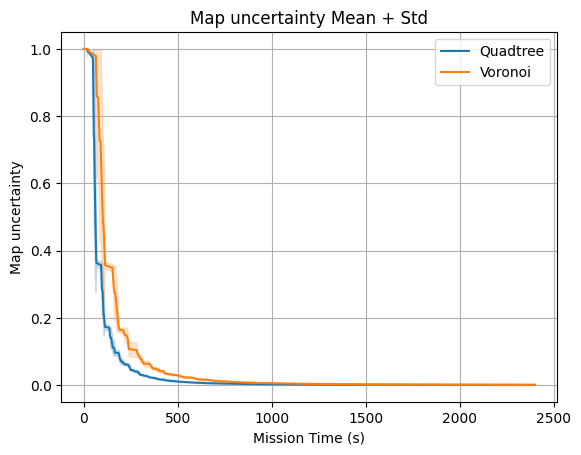}
         \label{fig:planner_map_uncertainty}
     \end{subfigure}
    \caption{Comparison of mission performance metrics over time for the IPP system using Quadtree and Voronoi spatial representations for candidate view generation.}
    \label{fig:planner_metrics}
\end{figure}

Figure~\ref{fig:planner_metrics} shows how path planning metrics change for each representation over time. Mean and standard deviation values are computed by averaging over multiple missions with varying initial UAV positions. The primary metric of interest is weed coverage, as it directly reflects performance in real-world scenarios where the objective is rapid identification of high-interest regions. We can see that the quadtree-based planner achieves a faster initial increase in weed coverage during the early stages of the mission. However, the Voronoi-based approach converges halfway through the mission and, in some cases, surpasses the quadtree in coverage efficiency, as indicated by the larger variance associated with the Voronoi representation. Both planners achieve nearly complete weed coverage by the conclusion of the 40-minute flight budget.

A similar trend is observed in total map coverage. The quadtree representation shows a more rapid initial expansion, whereas the Voronoi representation eventually overtakes it toward the latter stages of the mission.

RMSE quantifies how well the trained GP models the ground-truth orthomosaic. The quadtree representation initially reduces RMSE more rapidly with greater variance than the Voronoi representation. The Voronoi approach temporarily achieves superior performance before the quadtree ultimately produces a lower RMSE map by the mission’s end.

Map uncertainty measures how the mean of the covariance map changes over time. The overall uncertainty decreases sharply for both representations, following nearly identical trajectories. However, the quadtree representation achieves a marginally faster reduction in map uncertainty compared to the Voronoi representation.

\paragraph{Qualitative Results}

\begin{figure}[ht]
     \centering
     \begin{subfigure}[t]{0.0892\columnwidth}
         \centering
         \includegraphics[width=\columnwidth]{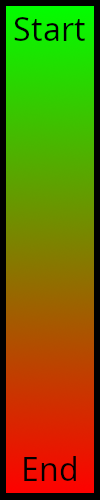}
     \end{subfigure}
     \begin{subfigure}[t]{0.446\columnwidth}
         \centering
         \includegraphics[width=\columnwidth]{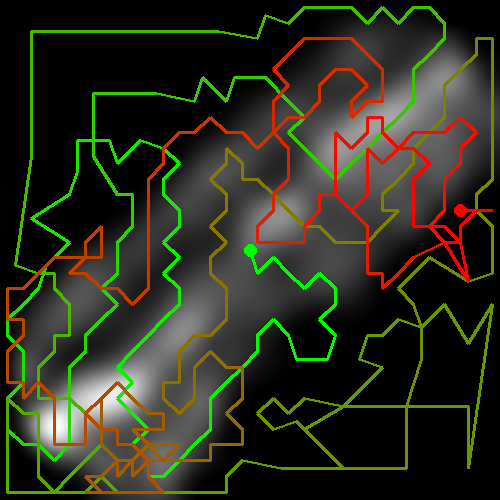}
         \caption{Quadtree}
         \label{fig:planner_quad_path}
     \end{subfigure}
     \begin{subfigure}[t]{0.446\columnwidth}
         \centering
         \includegraphics[width=\columnwidth]{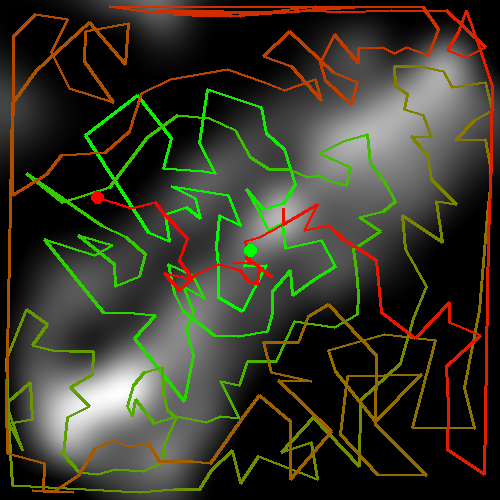}
         \caption{Voronoi}
         \label{fig:planner_voro_path}
     \end{subfigure}
    \caption{Comparison of UAV trajectories generated by the IPP framework using Quadtree and Voronoi representations for candidate view generation. The mission start and end locations are denoted by green and red markers, respectively.}
    \label{fig:planner_qual}
\end{figure}

Figure~\ref{fig:planner_qual} illustrates example trajectories generated by the IPP system using either quadtree or Voronoi for candidate view generation. These trajectories reveal systematic differences in exploration behaviour induced by the underlying spatial representation. The Voronoi-generated path is observed to remain in its initial location for a substantially longer duration than the quadtree-generated path. This pattern recurs throughout the mission, with the Voronoi path spending more time in newly visited areas before moving on and rarely revisiting them. In contrast, the quadtree path performs rapid initial passes through new regions and revisits them more frequently later in the mission.

Overall, the behaviour of the Voronoi path appears more exploitative, whereas the quadtree path exhibits more exploratory tendencies. The Voronoi representation provides greater flexibility in the geometric structure of its paths, allowing for circular or spiral trajectories, while the quadtree representation constrains motion direction and path shape, particularly over short distances or within local regions. This increased flexibility appears to disadvantage the Voronoi planner by promoting excessive time spent exploiting early locations. However, such flexibility may become beneficial in future work when combined with a more finely tuned planning strategy capable of leveraging the broader path adaptability inherent to the Voronoi representation. These results demonstrate that representation choice alone can bias the planner toward exploratory or exploitative behaviour.

\paragraph{Computational Performance}

\begin{figure}[t]
     \centering
     \begin{subfigure}[t]{0.48\columnwidth}
         \centering
         \includegraphics[width=\columnwidth]{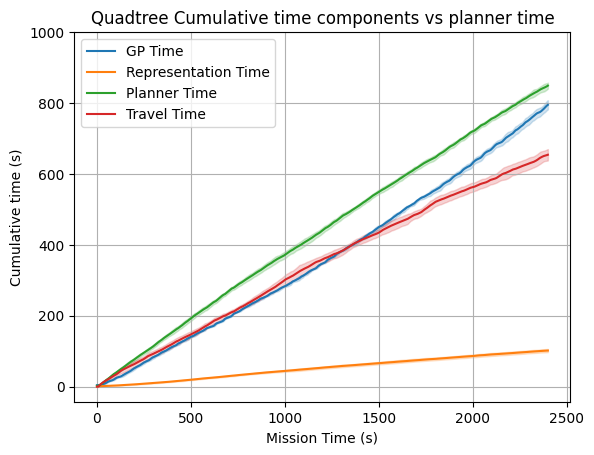}
         \caption{Quadtree}
         \label{fig:planner_quad_time}
     \end{subfigure}
     \begin{subfigure}[t]{0.48\columnwidth}
         \centering
         \includegraphics[width=\columnwidth]{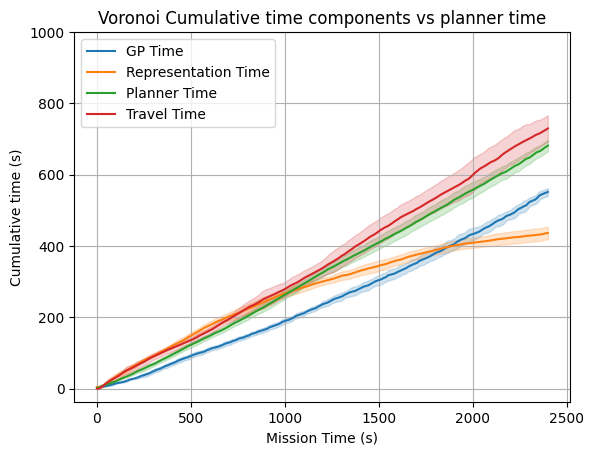}
         \caption{Voronoi}
         \label{fig:planner_voro_time}
     \end{subfigure}
    \caption{Breakdown of the time taken on different tasks for each representation. Graphs show time taken for: Generating / Training the GP, Generating the Discrete Representation, Planner choosing the next position and Travel Time assuming a constant speed of 2m/s.}
    \label{fig:planner_time}
\end{figure}

Figure~\ref{fig:planner_time} presents the computational performance of the different stages of the planner, highlighting how execution time varies according to the different discrete representations used. The most pronounced difference between the two representations is the substantially longer time taken to generate a Voronoi diagram compared to a quadtree. This is to be expected, as the Voronoi representation is much more complex

As the planning process operates under a fixed time budget, this computational overhead influences the allocation of time across other tasks. The results show that the Voronoi representation allocates a larger proportion of mission time to traversal, while the GP evaluation and planning stages require comparatively less time. Such behaviour may be advantageous in real-world deployments, where maximising the time spent on traversal directly increases data collection and spatial coverage within the UAV’s battery constraints. Minimising the computational time dedicated to GP updates and planning can enable more efficient utilisation of mission resources. 

\subsection{Discussion}

\paragraph{Representations}
As seen in Table \ref{tbl:representation_simularity_metrics}, the quadtree representation achieves the highest mean scores across all metrics and maps, especially in Hamming distance. However, BSP region-based and BSP LSE perform best for map 001’s MSE (high coverage, extreme fragmentation, large clusters). Hexagon and Voronoi maps excel in maps 003 and 004’s MSE (moderate coverage, many small clusters). BSP LSE also achieves the best SSIM for map 004 (many small clusters, largest individual cluster), while Voronoi produces the best SSIM for map 003.

\begin{table}[!t]
\centering
\caption{Recommended representation choice based on observed weed distribution conditions (lower score = better). Methods considered: Quadtree, Wedgelet, H3 (hex grid), BSP LSE, BSP Region, Voronoi.}
\small
\resizebox{\columnwidth}{!}{%
\begin{tabular}{p{0.28\columnwidth} p{0.68\columnwidth}}
\hline
\textbf{Condition} & \textbf{Suggested Method} \\
\hline\hline
High overall coverage        & \textbf{Quadtree or Wedgelet}: scores tend to \emph{decrease} as coverage increases; \emph{avoid BSP LSE} whose score rises with coverage. \\
Many smaller patches         & \textbf{Quadtree or Wedgelet}: particularly effective when patches are numerous and moderately sized. \\
One large (dominant) patch   & \textbf{BSP Region or Voronoi}: great improvements under dominance; Quadtree/Wedgelet largely insensitive. \\
Highly variable patch size   & \textbf{BSP Region or Voronoi}: handle heterogeneity well; BSP LSE shows only a weak tendency to worsen. \\
Many disjoint weed clusters  & \textbf{Quadtree or Wedgelet}; \emph{avoid H3}: H3’s score increases strongly with fragmentation. \\
Patches are large on average & \textbf{Quadtree or Wedgelet}; \emph{avoid BSP LSE}: BSP LSE’s score increases with average patch size. \\
\hline
\end{tabular}
}
\label{tab:method_guidelines}
\end{table}

Table \ref{tbl:representation_time_space_mean_std_of_mean} shows that quadtree and BSP region are the most computationally efficient methods, while the hexagon representation uses the least memory. BSP LSE requires over 2 minutes per representation~\cite{j_time_2012}, making it unsuitable for real-time UAV mapping, but it uses less than half the storage of the quadtree and wedgelet methods.

Correlation analysis points to clear operating regimes. Coverage-driven features are strongly positively associated with BSP LSE (indicating degradation as coverage intensifies), while quadtree/wedgelet tend to improve. Fields dominated by a single patch or exhibiting high patch-size variability favour BSP Region/Voronoi, whereas heavy fragmentation penalises H3. Larger average patches, as well as high coverage, generally favour quadtree/wedgelet and disfavour BSP LSE. Representation performance is strongly conditioned by the spatial properties of the environment. Table~\ref{tab:method_guidelines} summarises the recommended method per field condition, grounded in the signs and magnitudes of the correlations in Table~\ref{tab:pos_neg_corr}. In practice, this means preferring quadtree/wedgelet under high coverage or many/large patches, choosing BSP Region/Voronoi when dominance or heterogeneity is expected, avoiding H3 in highly fragmented settings, and avoiding BSP LSE when coverage or patch size increases.

\paragraph{Path Planner}

The experimental results demonstrate that the choice of spatial discretisation is a primary determinant of the emergent exploration dynamics and mission efficiency. While both representations achieved near-complete weed and map coverage within the 40-minute constraint, they exhibited distinct operational characteristics. The quadtree-based planner prioritised rapid, large-scale coverage in the early mission stages, reflecting a global exploration bias facilitated by its hierarchical structure. Conversely, the Voronoi-based planner adopted a more exploitative strategy, allocating extended periods to refining local regions. This led to slower initial coverage but allowed the Voronoi approach to occasionally surpass the quadtree in map fidelity during the mid-mission phase.

These behavioural differences directly impacted mapping accuracy. The quadtree representation facilitated a faster reduction in RMSE during early exploration through an extensive initial survey of the environment, allowing for the rapid identification of high-value areas. In contrast, the Voronoi approach achieved lower local error peaks by densely sampling clusters before transitioning to adjacent regions. Notably, while the quadtree ultimately converged to superior final accuracy, map uncertainty decreased at nearly identical rates for both methods, suggesting that the underlying IPP utility function maintains consistent exploration pressure regardless of the discretisation geometry.

Analysis of the resulting trajectories highlights the influence of the representation on the physical movement of the UAV. The axis-aligned nature of the quadtree encouraged a structured, grid-like exploration pattern with frequent revisits to central nodes. In contrast, the Voronoi planner produced smoother, curvilinear trajectories and longer dwell times in specific sub-regions. This spatial persistence in the Voronoi planner suggests an inherent adaptivity to the data distribution, though it occasionally resulted in local oversampling.

From a computational perspective, the results reveal an interplay between computational latency and operational execution. Generating the Voronoi discretisation incurred a higher initial overhead; however, once initialised, it permitted the UAV to allocate a greater proportion of the mission budget to traversal rather than frequent re-computation. This relationship is critical for real-world deployments, where the objective is to maximise the ratio of active data collection to computational downtime to ensure that the finite flight duration is utilised effectively.

\section{Conclusion}
\label{sec:conclusion}

This manuscript evaluated discrete spatial representations of GP maps as a key design choice in UAV-based field exploration. Our experimental results demonstrate that these alternatives can provide more accurate representations for specific weed distributions depending on the number, size, variation, and coverage of weeds compared to traditional grid maps and quadtrees while maintaining strong compression rates.

In addition, an online information-gain-based path planning framework was introduced to assess how different discrete GP representations influence mission-level performance for UAV-based weed mapping. By comparing quadtree and Voronoi-based discretisations within a fixed flight-time budget, the study showed that representation choice not only affects reconstruction accuracy but also the rate of weed coverage, map coverage, uncertainty reduction, and computational load allocation between planning and traversal.

Our results suggest that discretisation should be considered jointly with the planner and probabilistic model, rather than selected a priori. We provide a guideline through spatial analysis of what representations should be investigated, depending on the distribution of the underlying data. In general, for large dominant patches, BSP Region or Voronoi representations should be investigated for use as they are the most efficient for this distribution pattern; however, for distributions comprised of many dispersed regions, quadtrees provide the most benefit. 

Future work will extend this evaluation to more diverse datasets and incorporate planners specifically optimised for the respective spatial representations, rather than relying on general-purpose configurations. Additionally, real-world field tests are planned, alongside integrating altitude control to the IPP system, allowing dynamically adaptive measurement detail based on the expected information gain within regions of a field.

\subsection*{Declaration of generative AI and AI-assisted technologies in the manuscript preparation process} During the preparation of this work the author(s) used Chat-GPT and Perplexity for research and proofreading. After using this tool/service, the author(s) reviewed and edited the content as needed and take(s) full responsibility for the content of the published article.

\subsubsection*{Acknowledgements} This work was supported by Engineering and Physical Sciences Research Council, UK Projects GAIA (Ref: EP/Y003438/1) and AgriFoRwArdS (Ref: EP/S023917/1).

\bibliographystyle{elsarticle-num} 
\bibliography{glorified,new}

\end{document}